\documentclass[10pt,journal,compsoc]{IEEEtran}
\usepackage{booktabs} 
\usepackage{graphicx}
\usepackage{subfigure}
\usepackage{float}
\usepackage{color}
\usepackage[noend]{algpseudocode}
\usepackage{algorithmicx,algorithm}
\usepackage{times}
\usepackage{epsfig}
\usepackage{amsmath}
\usepackage{amssymb}
\usepackage{enumitem}
\usepackage{verbatim}
\usepackage{cite}
\usepackage{amsfonts}
\usepackage{enumerate}
\usepackage{array}
\usepackage{comment}
\usepackage{multicol}
\usepackage{multirow}
\usepackage{multirow}
\usepackage{epsfig}
\usepackage{booktabs}
\usepackage{subfigure}
\usepackage{float}
\usepackage{graphicx}
\usepackage{amsmath,bm}
\usepackage{amssymb}
\usepackage{amsfonts}
\usepackage{amsthm}
\usepackage{graphicx,lipsum}
\usepackage{multirow}
\usepackage{hyperref}
\usepackage{bbding}
\usepackage{tabularx}

\ifCLASSINFOpdf
\else
\fi
\begin{document}

\title{Reinforcement Learning Based Multi-modal Feature Fusion Network for Novel Class Discovery}

\author{Qiang Li,
        Qiuyang Ma,
        Weizhi Nie*,
        Anan Liu,~\IEEEmembership{Senior Member}
        
        ~
\thanks{Qiang Li, and Qiuyang Ma are with the School of Microelectronics, Tianjin University. Weizhi Nie and Anan Liu are with the School of Electrical and Information Engineering, Tianjin University. }
\thanks{Weizhi Nie is the corresponding author(email: weizhinie@tju.edu.cn).}}

\markboth{Journal of \LaTeX\ Class Files,~Vol.~14, No.~8, August~2015}%
{Shell \MakeLowercase{\textit{et al.}}: Bare Demo of IEEEtran.cls for Computer Society Journals}

\IEEEtitleabstractindextext{%
\begin{abstract}
With the development of deep learning techniques, supervised learning has achieved performances surpassing those of humans. Researchers have designed numerous corresponding models for different data modalities, achieving excellent results in supervised tasks. However, with the exponential increase of data in multiple fields, the recognition and classification of unlabeled data have gradually become a hot topic. 
In this paper, we employed a Reinforcement Learning framework to simulate the cognitive processes of humans for effectively addressing novel class discovery in the Open-set domain. 
We deployed a Member-to-Leader Multi-Agent framework to extract and fuse features from multi-modal information, aiming to acquire a more comprehensive understanding of the feature space. Furthermore, this approach facilitated the incorporation of self-supervised learning to enhance model training. We employed a clustering method with varying constraint conditions, ranging from strict to loose, allowing for the generation of dependable labels for a subset of unlabeled data during the training phase. 
This iterative process is similar to human exploratory learning of unknown data.
These mechanisms collectively update the network parameters based on rewards received from environmental feedback. This process enables effective control over the extent of exploration learning, ensuring the accuracy of learning in unknown data categories.
We demonstrate the performance of our approach in both the 3D and 2D domains by employing the OS-MN40, OS-MN40-Miss, and Cifar10 datasets. Our approach achieves competitive competitive results. The source code can be found here: \url{https://github.com/AveryFallson/RMFFN}
\end{abstract}

}
\maketitle

\IEEEdisplaynontitleabstractindextext

\IEEEpeerreviewmaketitle

\section{Introduction}

\IEEEPARstart{I}{n} recent years, driven by the ongoing advancement of deep learning technology, it has demonstrated exceptional performance surpassing human capabilities in the realm of data classification and retrieval. Illustrating with a 3D model as an exemplar, within a labeled training dataset, prevailing methodologies\cite{achlioptas2018learning,huang2019texturenet,landrieu2018large,qi2017pointnet,wu20153d} proficiently extract data's intrinsic feature information, accomplishing classification or retrieval tasks for homogeneous data. However, the tangible world constitutes a dynamic and open milieu where novel categories of objects incessantly emerge. The remarkable efficacy of established techniques in data recognition tasks predominantly hinges on the meticulous extraction of discernible information from known sample sets, as well as the utilization of elucidative descriptors to attain high-precision data recognition endeavors. Consequently, this dynamic poses a formidable hurdle for network models trained on the original dataset when confronted with the exigencies of tasks involving previously unseen class samples, as juxtaposed against their performance under closed-set circumstances. Meanwhile, seasoned practitioners expend significant temporal and cognitive resources to meticulously annotate the ever-expanding data volume. As a corollary, the growing number of 3D models brings great challenges to data annotation.

\begin{figure}[t]
    \centering
    \includegraphics[scale=0.32]{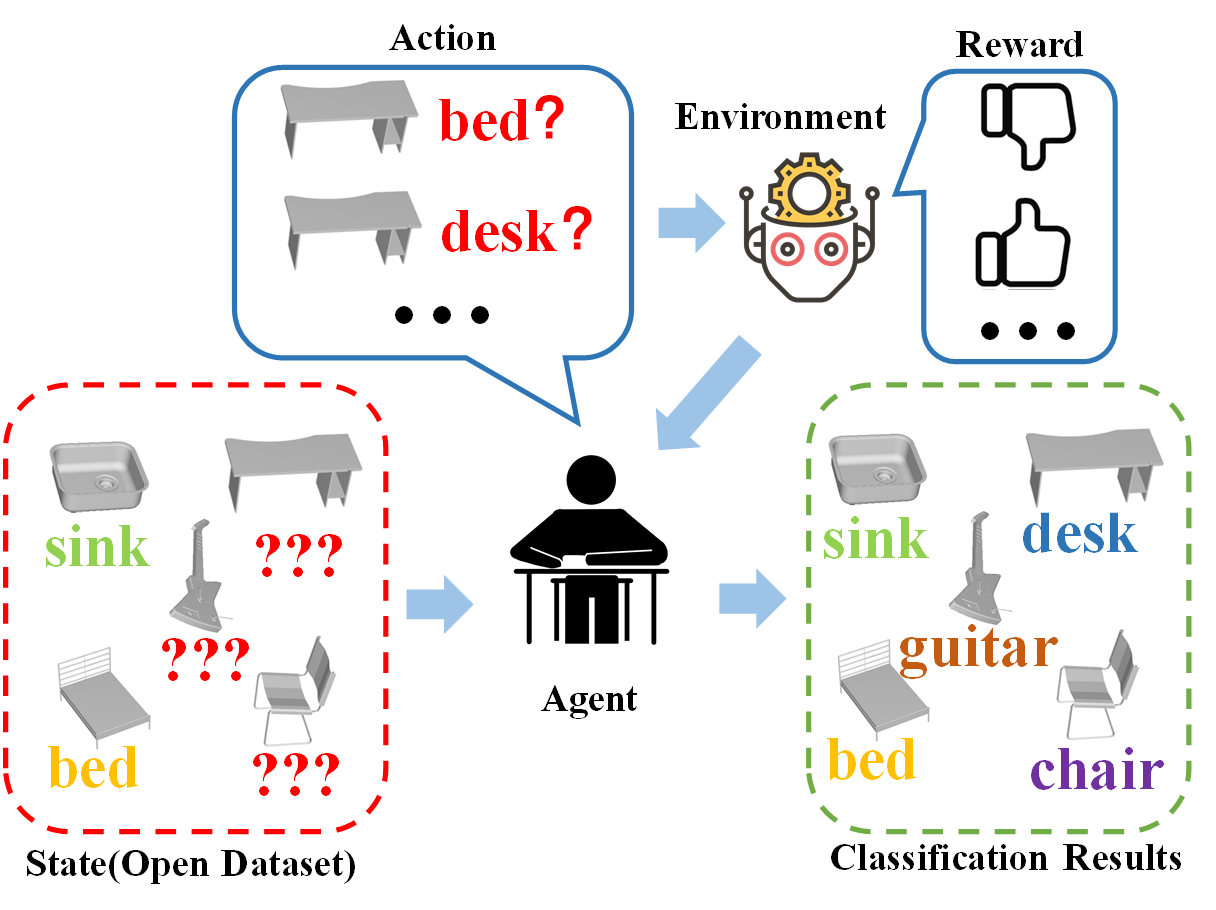}
    \caption{The motivation of our approach. Using a reinforcement learning framework, the agent is trained through rewards and penalties during the exploration process in an open-set situation to mimic the human learning process and accomplish the task of novel class discovery.}
    \label{fig.0}
\end{figure}

To alleviate these problems, Novel Class Discovery (\textbf{NCD}) \cite{bendale2015towards,xie2017aggregated,scheirer2012toward} has recently emerged as a practical solution. 
The work of Han et al.\cite{han2019learning} pioneered the formulation of the NCD problem by defining it as the task that aims to classify the samples of an unlabeled dataset into different classes by exploiting the knowledge of a set of labeled samples.
Novel class discovery refers to the process of identifying and categorizing previously unseen or unknown classes or categories in a given dataset. It involves discovering and classifying instances that do not belong to any pre-defined or known classes in the dataset.
The goal of NCD is to develop algorithms and techniques that can effectively identify and handle novel or unfamiliar classes. This task is challenging because it requires the ability to generalize knowledge from known classes to identify and categorize instances from new, unseen classes accurately.

\subsection{Motivation}

To our knowledge, the vast majority of existing NCD studies\cite{han2019learning,fini2021unified,xie2016unsupervised} focus on 2D images. 
We endeavor to investigate a NCD framework suitable for diverse data modalities. Drawing a compelling analogy to human learning processes, where existing knowledge guides the exploration of uncharted domains, we aspire to effectively simulate human behavioral patterns within the machine's learning paradigm. Similar to humans who assess their learning advancements through external feedback and adapt their strategies accordingly, we aim to integrate multi-modal information and emulate human learning behaviors. This holistic approach enhances our comprehension and resolution of the NCD challenge.

In this paper, we explore the new problem of NCD for multi-modal information.
The motivation of our approach is shown in Fig.\ref{fig.0}. To simulate the human learning process, we have adopted the fundamental framework of reinforcement learning\cite{sutton2018reinforcement}. 
In human cognition, a comprehensive evaluation often emerges from amalgamating insights from various perspectives. In line with this, multiple agents are employed within the reinforcement learning framework to process diverse data modalities. On one hand, the fusion of multi-modal features equips our model with a holistic grasp of labeled data. On the other, juxtaposing feature vectors from distinct models facilitates the generalization of unlabeled data into novel categories.

In addition, during the process of human learning, it is often possible to quickly grasp knowledge that is highly correlated with prior knowledge or relatively simple, whereas more complex knowledge requires greater effort. Similarly, following pre-training with labeled data, our machine learning model can obtain relatively reliable feature vectors for unlabeled data that exhibit high similarity to labeled data or possess lower complexity. By employing a clustering method with constraint conditions ranging from strict to loose, abbreviated as \textbf{STLClu}, we transform these unlabeled data into labeled data. As training progresses, the quantity of unlabeled data steadily decreases. Thus, this approach can effectively alleviate the training difficulty.

We aim for the network to automatically acquire knowledge from unlabeled data after grasping prior knowledge from labeled data, enabling effective completion of classification and retrieval tasks.
We build a Reinforcement Learning Based Multi-modal Feature Fusion Network (RMFFN), which updates the network parameters based on the rewards of environmental feedback by simulating the way humans learn novel things. Given a particular dataset, comprising a labeled dataset and an unlabeled dataset without shared categories, our approach, RMFFN, leverages supervised learning on the labeled dataset to obtain prior knowledge. This prior knowledge is then transferred to the unlabeled dataset, and the performance of agents is evaluated based on rewards provided by the environment. 
RMFFN exploits multi-modal data to acquire integrated global feature vectors. We adopt a hierarchical multi-agent framework, denoted as Member-to-Leader Multi-Agent (\textbf{MLMA}), for multi-modal data processing. The count of member agents aligns with the number of data modalities, with each agent dedicated to processing a specific modality, generating an associated feature vector as its action. Following this, the leader agent amalgamates member agents' actions and issues a fused global feature vector as the conclusive action furnished to the environment. Subsequently, the environment evaluates the fused global feature vector to provide agents with feedback in the form of rewards, prompting subsequent network parameter updates by the agents.
Additionally, we employ the STLClu strategy to consistently transform unlabeled data into labeled data throughout the training procedure. Moreover, we leverage multi-modal feature vectors for self-supervised learning to enhance the model's learning capacity.
To substantiate our approach, we utilized 3D model data encompassing four modalities: multi-view representation, point cloud, mesh, and voxel. Our method's performance was assessed on OS-MN40\cite{feng2022shrec} and OS-MN40-Miss\cite{feng2022shrec} datasets, yielding results that surpass the established baseline. Furthermore, we conducted a comprehensive ablation study to validate the effectiveness of each module in our method. Additionally, a transfer experiment on the Cifar10 \cite{krizhevsky2009learning} dataset was conducted to validate the generalizability of our approach.

\subsection{Contribution}
The key contributions of our work are followed as: 
\begin{itemize}
\item[$\bullet$]
We propose a novel multi-agent reinforcement learning framework that addresses the task of novel class discovery under open-set conditions by simulating the human learning process;
\item[$\bullet$]
We employ an MLMA mechanism, enabling us to effectively extract and fuse information across different modalities. Additionally, the self-supervised mechanism introduced for the multiple agents can significantly enhance the learning capacity of our model;
\item[$\bullet$]
We introduce the STLClu strategy, which enables us to gradually transform unlabeled data into labeled data during the training process to address the Novel Class Discovery problem.
\end{itemize}

The remainder of our paper is outlined below. In Section II, we present the related work. The proposed solution is presented in Section III. In Section IV, we present vital experiments. This section also includes a summary of the experimental results. Finally, in Section V, we make a conclusion for this work.

\section{Related work}

\subsection{Novel Class Discovery}

The real world is a vast and open domain where new entities continuously emerge, making the study of NCD a matter of great significance. In contrast to semi-supervised learning, the NCD approach assumes a disjoint relationship between labeled and unlabeled samples. 
Han et al.\cite{han2019learning} were pioneers in addressing the NCD problem in image classification. They train a classification model on a set of base classes and employ it as a feature extractor for the novel classes. Subsequently, they train a classifier using the pseudo-labels generated by the pre-trained model. This basic framework forms the foundation for much of the NCD research\cite{riz2023novel,fini2021unified,zhao2021novel}, i.e. using clustering method\cite{likas2003global,kumar2014canopy,murtagh2012algorithms,ester1996density} to generate pseudo-labels for unlabeled data.
Building upon this foundation, Fini et al.\cite{fini2021unified} explored the use of the Sinkhorn-Knopp algorithm\cite{knight2008sinkhorn} to generate pseudo-labels and proposed a Unified Objective for NCD. Their approach treats cluster pseudo-labels on par with ground truth labels, allowing a single Cross-Entropy (CE) loss to operate on both labeled and unlabeled sets. 
Drawing inspiration from this work, Riz et al.\cite{riz2023novel} tackled the problem of NCD in 3D semantic segmentation. They introduced NOPS, an approach that addresses NCD for point cloud segmentation by employing online clustering and leveraging uncertainty quantification to generate pseudo-labels for the novel points. 
Incorporating the idea of contrastive learning into the NCD problem, Zhong et al.\cite{zhong2021neighborhood} presented Neighborhood Contrastive Learning. Their method effectively harnesses the local neighborhood information in the embedding space, allowing for the utilization of knowledge from more positive samples and thereby improving clustering accuracy. 
Zhao et al.\cite{zhao2022novel} focused on segmenting unlabeled novel classes using prior knowledge obtained from labeled data of disjoint categories. They proposed a basic framework that leverages labeled base data and a saliency model to discover novel classes in unlabeled images through clustering. 
Shermin et al.\cite{9165955} introduced multiple classifiers within an adversarial framework to enhance domain adaptation in scenarios where target domain classes are incompletely known.
These works contribute to the field of NCD by exploring different aspects such as image classification, 3D semantic segmentation, contrastive learning, and leveraging prior knowledge. Their contributions enhance our understanding of NCD and provide valuable insights for future research.

\subsection{Reinforcement Learning}

\begin{figure*}
\centering
\includegraphics[scale=0.5]{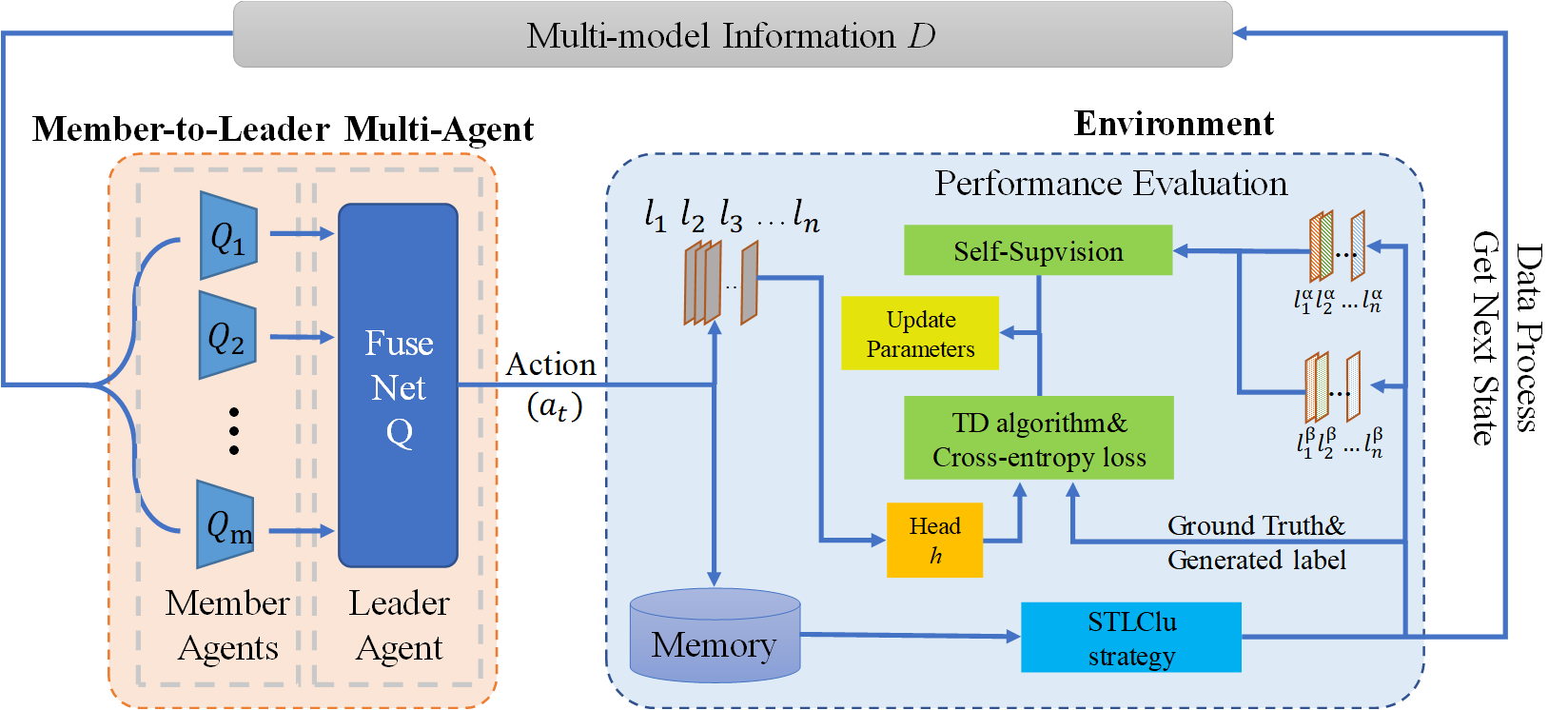}
\caption{The framework of our approach. It is based on the principles of Reinforcement Learning. The multi-modal information serves as the state.
Our member agents independently extract feature vectors from diverse modal data and transmit them to the leader agent for fusion. 
The fused global feature vector $l_i$ is generated by the leader agent and serves as the action $a_t$ for interacting with the environment, subsequently being stored in the environment's memory.
Upon the completion of an epoch, we apply the STLClu strategy to the actions associated with all stored unlabeled data, resulting in the generation of labels for a subset of these unlabeled instances. These labels are then applied in the subsequent epoch for loss function computation. Furthermore, we incorporate a self-supervised learning approach to facilitate the training of the model.}
\label{fig.1}
\end{figure*}

Reinforcement learning\cite{sutton2018reinforcement}, a subfield of machine learning, focuses on developing algorithms to enable agents to make optimal decisions in order to maximize cumulative reward signals within an environment. The primary objective is to cultivate agents capable of learning through trial and error, without explicit programming. In reinforcement learning, agents learn from interactions with their environment, receiving observations on the current state and selecting actions accordingly. Following each action, the environment transitions to a new state, providing a reward signal based on the action and resultant state. The agent's ultimate goal is to learn a policy that effectively maps observations to actions, thereby maximizing expected cumulative rewards over time. 
In the field of classification and segmentation, researchers have achieved promising outcomes by applying reinforcement learning methods\cite{zhao2016deep,mousavi2019multi,huang2019fully,liu20173dcnn,mou2021deep,nie2023knowledge}. 
For instance, Liao et al.\cite{liao2020iteratively} proposed a multi-agent strategy for 3D model segmentation, leveraging reinforcement learning principles. 
Additionally, Nie et al.\cite{nie20203d} introduced a novel reinforcement learning model for estimating 3D poses based on 2D images. These studies showcase the effectiveness of reinforcement learning in addressing classification and segmentation challenges.
Zhao et al.\cite{zhao2016deep} computed the information entropy to guide a reinforcement learning agent to achieve a better policy for image classification to select the key parts of an image.
Chen et al.\cite{9359523} proposed a drift-proof tracker with deep reinforcement learning that aims to improve tracking performance by counteracting drifts while maintaining its real-time advantage.
Chen et al.\cite{chen2018recurrent} proposed a recurrent attention reinforcement learning framework to iteratively discover a sequence of attentional and informative regions that are related to different semantic objects and further predict label scores conditioned on these regions.
Hafiz et al.\cite{hafiz2022image} used a Q-learning with an agent having two states and two to three actions to achieve an efficient and simple image recognition classification system.
Li et al.\cite{li2020deep} proposed a novel framework based on reinforcement learning for pre-selecting useful images for emotion classification.

\subsection{3D Models Classification and Retrieval}
As we mainly validated our approach using 3D model data, we now provide a brief overview of the research in the realm of 3D model classification and retrieval.
The number of different methods for 3D model recognition has increased considerably in recent years\cite{nie20163d,wang2019learning,wu20153d}. In supervised learning, satisfactory performance can be achieved in both single-modal and multi-modal scenarios.
Su et al.\cite{su2015mvcnn} attempted to recognize 3D shapes by utilizing rendered views of these shapes from 2D images. They introduced a novel CNN architecture that effectively integrates information from multiple views, resulting in a concise shape descriptor with improved recognition performance.
Qi et al. \cite{qi2017pointnet} developed a groundbreaking neural network architecture that directly processes point clouds, providing a unified framework for various applications such as object classification, part segmentation, and scene semantic parsing.
Ma et al.\cite{8490588} combined convolutional neural networks with long short-term memory to exploit the correlative information from multiple views.
Wang et al.\cite{wang2019dynamic} proposed a dynamic graph convolutional neural network (DGCNN) approach that captures local geometric structures and effectively aggregates local features for learning on point clouds.
Feng et al.\cite{feng2018meshnet} introduced MeshNet, a mesh neural network that learns 3D shape representations directly from mesh data.
Zhou et al.\cite{zhou2018voxelnet} introduced VoxelNet, a versatile 3D detection network that integrates feature extraction and bounding box prediction into a single-stage, end-to-end trainable deep network.
Han et al.\cite{han2018seqviews2seqlabels} proposed a method that utilizes recurrent neural networks (RNN) with attention mechanisms to aggregate sequential views of 3D shapes and learn global features.
Song et al.\cite{9165939} a universal cross-domain 3D model retrieval framework is proposed for utilizing the labeled 2D images or 3D models to manage unlabeled 3D models with no prior knowledge about label sets.
In addition to utilizing single-modal data as inputs to neural networks, many researchers have conducted extensive studies on leveraging the multi-modal information of 3D models.
Zadeh et al.\cite{zadeh2018memory} introduced a novel framework that combines memory networks and fusion networks to capture sequential information from multiple views.
Zhang et al.\cite{10159496} proposed PointMCD, a unified multi-view cross-modal distillation architecture, including a pre-trained deep image encoder as the teacher and a deep point encoder as the student.
Hegde et al.\cite{hegde2016fusionnet} proposed that FusionNet combines multiple data representations to capture complementary information and improve classification performance.
Nie et al.\cite{nie2020mmfn} proposed an innovative multi-modal information fusion network for 3D shape recognition, leveraging the correlations between different modalities to generate a fused descriptor.

\section{Method}\label{sec:method}

In this section, we first introduce an overall framework of our approach and then illustrate the details of our approach in the next subsections.

\subsection{Overview}
In general, our approach is established on the concept of Reinforcement Learning. The classification process of multi-modal data can be viewed as a Markov Decision Process.
Given a batch of data, our agents determine the class to which each object belongs. Before this step, the agent acquires some prior knowledge from the labeled data. However, as the class of unlabeled data is unknown, the agents need to explore and discover the underlying patterns within the data. To provide feedback on the agents' actions, we construct a specialized environment. The framework is illustrated in Fig.\ref{fig.1}. The specific definitions and details are outlined below.

\begin{itemize}
\item[$\bullet$]
\textbf{State:} We define our dataset as $D$, which comprises both labeled sets $D_l$ and unlabeled sets $D_u$. Each dataset includes individual objects represented by $x_i \in D$. The dataset consists of several distinct modalities of information, represented by $D^1, D^2,\ldots, D^m$, where $m$ represents the number of modalities, i.e.,$D=\{D^1_{l},D^1_{u},D^2_{l},D^2_{u},\ldots, D^m_{l},D^m_{u},\}$, $x_i=\{x^1_{i},x^2_{i},\ldots, x^m_{i}\}$. Assuming that the number of objects in a batch is $n$, then the state $s_t$ can be represented as $s_t=\{ x_1,x_2,\ldots, x_n\}$. We will provide a detailed explanation of the dataset configurations and the method used for extracting multi-modal information in Section IV. 

\item[$\bullet$]
\textbf{Agent:} 
In our deployment, a collective of $m+1$ agents was established, encompassing $m$ member agents $(Q^1, Q^2,\ldots, Q^m)$ and one leader agent $(Q)$. The member agents are predominantly characterized by encoders with the ability to extract feature information$(l^1_{i},l^2_{i},\ldots, l^m_{i})$ aligned with their specific modalities, while the leader agent integrates a feature fusion network. This configuration emulates the cooperative division of tasks observed in human endeavors, wherein member agents contribute object cognition from diverse viewpoints, culminating in the leader agent's ultimate decision following a thorough evaluation.

\item[$\bullet$]
\textbf{Action:} 
Agent exhibits distinct actions in response to various states. 
The leader agent considers the feature information provided by the member agents. We utilize the fused global feature vector, represented as $l_i$ and produced by the leader agent, as the action $a_t$ to interact with the environment.
Our objective is for the agents to demonstrate actions that exhibit maximum dissimilarity when confronted with different categories of objects while maintaining a relatively high level of similarity in actions when faced with objects from the same category.

\item[$\bullet$]
\textbf{Environment:} 
In reinforcement learning, the environment serves as the entity that interacts with the agents, providing both the subsequent state and corresponding feedback based on the agent's actions.
Our environment, denoted as $E$, maintains a memory of all the agents' actions, utilizing them to generate reliable labels for unlabeled data. 
During the agent-environment interaction, our environment randomly selects a batch of data from the database. These data instances constitute the subsequent state $s_{t+1}$. Within the environment, we have established a reward mechanism that facilitates the provision of rewards to the agent. 

\item[$\bullet$]
\textbf{Reward:} We refer to the values that the environment feeds back to the agent for tuning the parameters of the encoder as reward $R_t$.
For each object classification result given by the agent, we compare it with the ground truth or generated label, where $R_t$ equals 1 if they agree, and 0 otherwise. In reinforcement learning, a discount factor $\gamma\in[0,1]$ is usually used to relate the importance of future rewards. Then the accumulative expected reward $V_t$ of the state $s_t$ can be expressed as $V_t=\sum_{j=0}^{\inf} \gamma^j\cdot R_{t+j}$
\end{itemize}

Our objective, based on the aforementioned definitions, is to train our agents capable of addressing the NCD problem for multi-modal information. Our agents aim to accurately recognize known categories of objects and subsequently classify a portion of unlabeled objects with discernible features based on their acquired prior knowledge.
As the agents gain proficiency in handling this subset of unlabeled objects, the environment proceeds to select another subset with distinctive features from the remaining unlabeled data for further training. Through this iterative process, the agent continuously enhances its knowledge, eventually achieving the classification of all unlabeled objects. To simulate this procedure, we construct a reinforcement learning framework where the goal is to maximize the expected cumulative reward $Q^{\ast}(s,a)$. Specifically, we aim to estimate the expected reward associated with taking action $a$ in state $s$, according to a given classification policy. This can be mathematically expressed as follows:

\begin{equation}
\begin{split}
 Q^{\ast}(s,a)&=\mathbb{E}[R_{t+1}+{\gamma}Q(s_{t+1},a_{t+1})]\\
 &=R_s^a+\gamma\sum_{s^\prime \in{S}} P^a_{{ss}^\prime} V(s^\prime),
\end{split}
\end{equation}

\begin{equation}
    V(s)=\sum_{a\in{A}}P_{sa}Q(s,a),
\end{equation}
where $P^a_{{ss}^\prime}$ denotes the probability of obtaining a particular state $s^\prime$ after taking action $a$ in state $s$. $P_{sa}$ denotes the the probability of taking a particular action $a$ in state $s$ under a particular classification policy.

Our goal is to find an optimal classification policy to solve the NCD problem. The optimal policy corresponds to the highest value of $V^{\ast}(s)$ and $Q^{\ast}(s,a)$, formulated as:

\begin{equation}
V^{\ast}(s)=max_{a}Q^{\ast}(s,a),    
\end{equation}

\begin{equation}
Q^{\ast}(s,a)=R_s^a+\gamma\sum_{s^\prime \in{S}}P^a_{{ss}^\prime} V^{\ast}(s^\prime). 
\end{equation}

Since the transition probability distribution $P$ is hard to obtain, we use the sampling method to address this issue in practice. By combining Equation (3) and (4), we can obtain the following expression:

\begin{equation}
\begin{split}
&Q(s,a)\approx R_s^a+\gamma max_{a^\prime}Q({s^\prime},{a^\prime})\\
&\approx(1-\alpha)Q(s,a)+\alpha[R_s^a+\gamma max_{a^\prime}Q({s^\prime},{a^\prime})]\\
&=Q(s,a)+\alpha[R_s^a+\gamma max_{a^\prime}Q({s^\prime},{a^\prime})-Q(s,a)]\\
&=Q(s,a)+\alpha{\delta_{TD}}.    
\end{split}
\end{equation}

Here, we apply the vanilla temporal difference(TD) algorithm \cite{watkins1989learning} to train the DQN and define the loss $\mathcal{L}_{TD}$ for the RL part based on the one-step TD error. For the discount factor $\gamma$ in $V_t$, since all of the objects are independently and identically distributed, we assume that the agents could only focus on the current reward, i.e. $\gamma=0$. Therefore, $\mathcal{L}_{TD}$ can be expressed as follow:

\begin{equation}
\mathcal{L}_{TD}={\frac{1}{2}}(Q(s,a)-R_t)^2.
\end{equation}

In our framework, the state $s_t$ represents a batch of multi-modal information. The agents extract fused multimodal features from the objects and output them to the environment as the action $a_t$. To facilitate training, we incorporate additional loss functions alongside the loss function $\mathcal{L}_{TD}$. The environment $E$ encompasses the ground truth $y^t$ of labeled data and the generated labels $y^p$ of unlabeled data. To address the optimization problem, we employ the simple cross-entropy loss for this component. 
Moreover, the MLMA strategy we use allows a self-supervised approach to be deployed to help the model train. Consequently, the final loss function can be formulated as follows: 

\begin{equation}
\mathcal{L}=\mathcal{L}_{TD}+\mathcal{L}_{CE}+\mathcal{L}_{SS}, 
\end{equation}
where $\mathcal{L}_{CE}$ denotes the cross-entropy loss and $\mathcal{L}_{SS}$ denotes the loss function associated with the self-supervised method. We will detail these losses in the later implementation stage.

\subsection{STLClu Strategy}

\begin{figure}[H]
\centering
\includegraphics[scale=0.37]{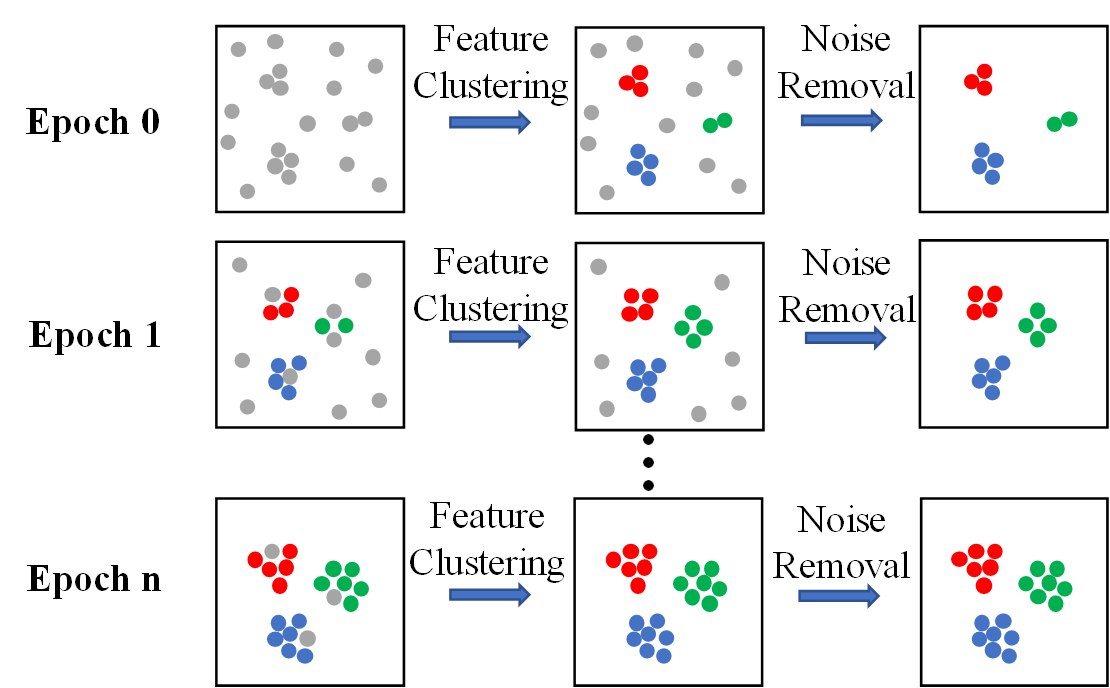}
\caption{The framework of our STLClu strategy. It begins by applying strict constraints, resulting in a limited but dependable set of clustered outcomes. Subsequently, looser constraints are adopted to reduce the amount of noise points. Through iterative repetition of this process, labels for previously unlabeled data are generated.}
\label{figclu}
\end{figure}

The process of human learning often involves deriving new knowledge from previously acquired understanding. When confronted with novel objects, experienced individuals can adeptly categorize them based on salient features. To tackle classification and retrieval tasks in an open-set scenario, we employ a clustering method with constraint conditions ranging from strict to loose, i.e., the STLClu strategy. This enables machines to emulate the process of human learning. The detailed process of STLClu is illustrated in Fig.\ref{figclu}.

After pre-training with labeled data, our model acquires the ability to recognize certain features. Consequently, feature vectors of unlabeled objects that share these recognized features or exhibit relative simplicity tend to show higher similarity, resulting in shorter distances within the mathematical space. Thus, we opt for the density-based clustering method DBSCAN\cite{ester1996density} as the foundational approach for STLClu.
By performing clustering on the labeled data and comparing it against ground truth, we extract a set of fundamental DBSCAN hyperparameters. Following meticulous calibration, we apply clustering to the feature vectors of unlabeled data, eliminating noise points and assigning the remaining clustering results as generated labels corresponding to the data.
In the subsequent iteration, we relax the clustering constraints, entailing larger radius parameters and smaller density thresholds. This adjustment amalgamates some noise points from the previous clustering round into their respective categories. This iterative process progressively transforms our unlabeled data into labeled data.

We have established a memory within the environment, which comprises two main components: firstly, the actions of agents during each epoch, and secondly, the labels corresponding to all data, encompassing both the ground truth for labeled data and the generated labels for unlabeled data. At the end of each epoch, the STLClu strategy is applied to all actions stored in the memory, resulting in the generation of labels for non-noise data points. It is worth noting that once unlabeled data acquires a generated label, that label remains fixed, even if the data point is identified as noise in the following epoch. Subsequently, the actions stored in the memory are cleared, marking the beginning of the next cycle.

We employ a head $h$ to process each $l_{i}$ into probability vector $\hat{y}{\in}\mathbb{R}^c$, $c$ is the number of categories in the whole dataset. Thus, the loss function $\mathcal{L}_{CE}$ can be written as:

\begin{equation}
\mathcal{L}_{CE}=\sum_{i=1}^n{-u_i{\cdot}y^t_ilog(\hat{y}_i)+(u_i-1){\cdot}y^p_ilog(\hat{y_i})}, 
\end{equation}
where $u_i$ equals to 1 when $x_i\in D_l$ and equals to 0 otherwise. $y^t_i$ represents the ground truth label of labeled data, while $y^p_i$ signifies the generated label for unlabeled data.

\subsection{Member-to-Leader Multi-Agent}\label{sec:env}

Human beings frequently engage in the exploration of novel concepts through collaborative efforts. Every team member contributes a distinct viewpoint when examining subjects, potentially encompassing imperceptible insights. The leader of the team will compile individual observations and undertake a thorough analysis, culminating in collaborative decisions for the entire team. Drawing inspiration from this notion, machine-based object classification similarly benefits from considering the multi-modal information of objects for more comprehensive and accurate feature extraction.

\begin{figure}[H]
\centering
\includegraphics[scale=0.36]{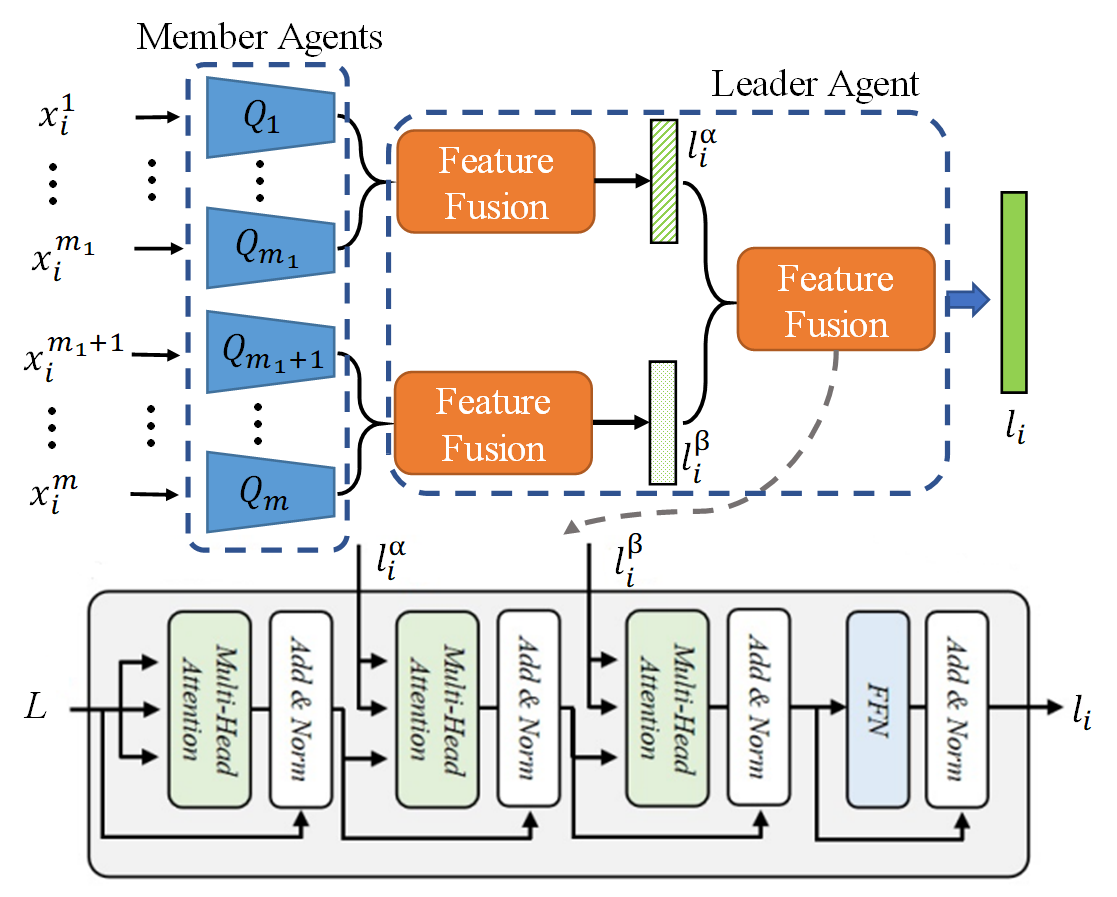}
\caption{The framework of our Member-to-Leader Multi-Agent. We present the network architecture for the final fusion step, $L$ represents the label embedding calculated by \cite{liu2021query2label}. When fusing the feature vectors provided by member agents, if the number of vectors to be fused exceeds two, we only need to add a corresponding number of multi-head self-attention layers.}
\label{figagent}
\end{figure}

Building upon this foundation, we have replicated the collaborative division of labor observed in human endeavors by establishing a Member-to-Leader Multi-Agent framework. More details are presented in Fig.\ref{figagent}.

For a given object denoted as $x_i$, comprising data from $m$ modalities, each member agent autonomously analyzes a specific modality and generates a feature vector $l_i^j$. Following this, our leader agent aggregates these feature vectors and executes feature fusion, yielding a global feature vector $l_i$, which serves as the action to interact within the environment.

Concerning the selection of feature extraction networks for member agents, we employ established classical network architectures tailored to each modality.
As for the leader agent, we introduce a multi-head self-attention\cite{vaswani2017attention} mechanism to facilitate feature fusion. Notably, the leader agent divides collected feature vectors into two groups, each encompassing $m1$ feature vectors. Should the number of modalities be odd, a zero vector is appended. 
Subsequently, each group uses a self-attention mechanism to fuse feature vectors, resulting in two fused feature vectors, $l_i^\alpha$ and $l_i^\beta$. As the fused feature vectors correspond to various modalities of the same object, a high level of similarity is expected. This motivates the use of these fused feature vectors to employ a self-supervised learning approach, effectively aiding the model's training process.The loss function $\mathcal{L}_{SS}$ can be defined as follows:

\begin{equation}
\mathcal{L}_{SS}=-{\frac{1}{n}}{\sum_{i=1}^n} log\frac{e^{cos(l_i^\alpha,l_i^\beta)/\tau}}{{\sum_{j=1}^n {e^{cos(l_i^\alpha,l_j^\beta)/\tau}}}},
\end{equation}
where $\tau$ is a temperature parameter. It is set to ${\tau}=2$ in our method. $cos(x,y)$ denotes the cosine similarity of $x$ and $y$.

Afterward, the leader agent will once again employ the self-attention mechanism to fuse vectors $l_i^\alpha$ and $l_i^\beta$, yielding the final fused global feature vector $l_i$.

At the beginning of training, our agents lacked knowledge of the unlabeled data and randomly assigned categories to these data. To emulate the human learning process, we employ the $\varepsilon$-greedy policy, which encourages exploration during training. This policy entails exploring with a probability of $\varepsilon\in[0,1]$ and utilizing learned knowledge with a probability of $(1-\varepsilon)$, allowing for the selection of non-optimal cases. In this setting, our goal is to explore non-optimal actions to make full use of our learned prior knowledge. Initially, as the agents cannot effectively identify the unlabeled data, we assign a relatively large value to $\varepsilon$ to encourage more exploration. As training progresses, the agents gradually acquire knowledge of the unlabeled data. At this stage, we shift the agents' focus towards exploitation and reduce the frequency of random exploration. 
Therefore, the expression for $\varepsilon$ is given by:

\begin{equation}
  \varepsilon=max\{ \varepsilon_{min},1-\frac{(1-\varepsilon_{min})*step}{total}\},  
\end{equation}
where $total$ denotes the number of iterations, and $step$ denotes the number of iterations that have been completed. $\varepsilon_{min}$ is a hyperparameter, we adjust its value to ensure that the model maintains a certain level of exploration capability even in the later stages of training.


\section{Experiment}
\subsection{Experiment Setting}
\subsubsection{Dataset}

We used multi-modal data from 3D models to demonstrate the effectiveness of the proposed method on retrieval and classification tasks. Our experiments were performed on the OS-MN40\cite{feng2022shrec} and OS-MN40-Miss\cite{feng2022shrec} datasets.
Furthermore, to validate the cross-domain generalization of the framework proposed by our method, we conducted a transfer experiment on the well-established NCD benchmark following the methodology of \cite{han2020automatically} based on Cifar10\cite{krizhevsky2009learning}.

\begin{figure}[H]
\centering
\includegraphics[scale=0.5]{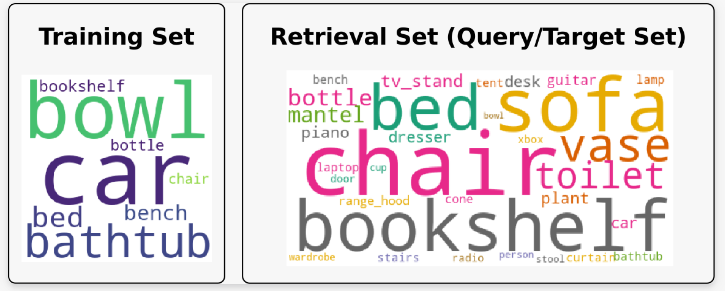}
\caption{Category information of training set and retrieval set of OS-MN40.}
\label{fig.3}
\end{figure}

OS-MN40 is an Open-set 3D object retrieval dataset, represented with multi-modality and multi-resolution. Most objects in OS-MN40 are selected from the ModelNet40\cite{wu20153d} dataset. OS-MN40 consists of 12309 objects from 40 classes. Eight classes are selected for training while the other 32 classes are selected for retrieval. The detailed category information of the training set and retrieval set is shown in Fig.\ref{fig.3}. For each object, it contains four modalities (Point cloud, Voxel, Multi-view, and Mesh). Note that the categories in training and retrieval are not shared.

OS-MN40-Miss is another version of OS-MN40. It is constructed by random drop arbitrary modality with probability 0.4 for each object. OS-MN40-Miss is collected towards the modality missing problem. For each category in the retrieval set, 30 samples are used for query, and the rest are used for target.

CIFAR-10 is a widely used dataset in the field of machine learning and computer vision. It consists of 60,000 32x32 color images spread across 10 different classes, with each class containing 6,000 images. Each class contains a balanced number of images, and the dataset is split into two parts: a training set of 50,000 images and a test set of 10,000 images. In this dataset, the images of the first five categories (airplane, automobile, bird, cat, deer) are treated as the labeled set, while the remaining five categories (dog, frog, horse, ship, truck) constitute the unlabeled set. To sum up, these three datasets can be summarized in Tab.\ref{Tab.1}.

\begin{table}[htpb]
\centering
\caption{The statistics of the dataset: train, target, and query.} 
\label{Tab.1}
\resizebox{\linewidth}{!}{
\begin{tabular}{ccccccc}
\hline
        &\multicolumn{3}{c}{OS-MN40/OS-MN40-Miss} &\multicolumn{3}{c}{Cifar10} \\ \cmidrule(r){2-4} \cmidrule(r){5-7}
        & Train  & Target & Query & Train  & Target & Query \\
\hline
Number of Object & 2822 & 8527 & 960 &  25000 & 25000& 10000\\ 
\hline
\end{tabular}}
\end{table}

\subsubsection{Evaluation Criteria}

In our experiments, we evaluate the classification performance of the proposed method using classification accuracy. For the retrieval task, we utilize multiple evaluation metrics\cite{liu2017view}, including Precision-Recall Curve, NN, mAP, NDCG, and ANMRR. In this context, a lower value of ANMRR indicates better retrieval performance, while higher values indicate better performance for the other metrics.

\begin{itemize}
\item[$\bullet$]The Precision-Recall Curve (PR-Curve) can comprehensively demonstrate retrieval performance. To change the threshold, which is used to distinguish the irrelevance and correlation in the model retrieval, it can jointly consider the accuracy and recall metrics.
\item[$\bullet$]The Nearest Neighbor (NN) indicates the percentage of the closest matching models.
\item[$\bullet$]The Mean Average Precision (mAP) is a ranking measure, which can solve the single point value limitation of Precision, Recall, and F-measure
\item[$\bullet$]The Normalized Discounted Cumulative Gain (NDCG) is a statistic that gives more attention to the top matching results. Here, we used the first 100 items from the retrieval results.
\item[$\bullet$]The Average Normalized Modied Retrieval Rank (ANMRR) presents the ranking performance of the ranking list, which takes the ranking information of the relevant models into account from the most frequently retrieved models.
\end{itemize}

\subsubsection{Implementation Details}

For 3D models, we collect the four different modal information by Blender. The multi-view image size is 224 × 224 × 3, and the point cloud data of each 3D model is represented by the three-dimensional coordinates of 1024 points. The volume of the voxel data is represented by 32 × 32 × 32 regions, and the mesh data is composed of 500 triangular surfaces. We utilize MVCNN\cite{su2015mvcnn} for extracting features from multi-view data, employ PointNet\cite{qi2017pointnet} for processing point clouds, and utilize MeshNet\cite{feng2018meshnet} and VoxelNet\cite{zhou2018voxelnet} for handling meshes and voxels, respectively.
In order to ensure the unity of feature vector dimensions, a fully connected layer is added after each branch of the feature extraction network so that the output feature vector dimension of each branch is 512. The added fully connected layer will be trained in conjunction with other networks during training. 

As for 2D images, we employ data augmentation techniques to obtain two distinct representations for each image, which are employed as the multi-modal information of the images. These augmentation methods include ResizedCrop, HorizontalFlip, ColorJitter, and Grayscale, randomly applied with specified probabilities. We utilize the classic ResNet-18\cite{he2016deep} as the encoder. Both of these dissimilar representations are fed into a shared encoder, yielding two positively correlated feature vectors. These feature vectors are then utilized for self-supervised learning and subsequently concatenated before being input into a multi-layer perceptron to generate the classification results.

In the method proposed in this paper, the Adam optimizer with a weight decay of 0.0001 is used to optimize the network parameters in the training stage. The initial learning rate is set to 0.001, and the learning rate decreases linearly with the number of iterations.
Our model initially initialized its parameters randomly, followed by undergoing 10 epochs of pre-training using labeled data. For the 3D models, the pre-trained model underwent an additional 40 epochs of training using the complete dataset with a minibatch size of 8. Each epoch required approximately 30 minutes to complete. Conversely, for the 2D images, the pre-trained model underwent an extended training phase of 200 epochs, utilizing the complete dataset with a minibatch size of 128. The completion time for each epoch was approximately 2 minutes. 
The computer used for the experiments was equipped with two NVIDIA 1080Ti GPUs, 32 GB RAM, and an Intel R ©Xeon(R) E5-2609 V4 1.70 GHz × 8 CPU. We utilize the PyTorch platform to make all experiments.

\subsection{Comparison With The State-of-art Methods}

In order to evaluate the retrieval performance of our method, we choose some state-of-the-art methods\cite{feng2022shrec} as a comparison. All the competing methods take the same experimental settings as the proposed method, and we choose the best results for comparison. Since different methods used different modalities, we show the information of the modality used by each method in Tab.\ref{Tab.2}. The experimental results of CAD-to-CAD 3D Object Retrieval are shown in Tab.\ref{Tab.3}.

As for OS-MN40, methods that use all four kinds of modalities outperform those that use only part of a single modality. The methods using partial modalities have an average of 0.3995, 0.8375, 0.5617, and 0.5901 on the mAP, NN, NDCG, and ANMRR metrics. The method using all four modalities has an average value of 0.5535, 0.8913, 0.6616, and 0.4762 on the mAP, NN, NDCG, and ANMRR metrics. It can be found that the use of multi-modal information can effectively improve the performance of the network.
Among the three methods using four modalities, both HCMUS and our method learn from unlabeled data during the training stage. Compared to the NUC\_AICV method that only trains the network model on the labeled dataset, the two methods that learn from unlabeled data show a $16.78\%$-$23.77\%$ improvement in mAP metric. This indicates that learning new class data through unsupervised methods can effectively enhance the performance of network models on an open-set. Compared to the HCMUS method, which also considers unlabeled data information, our method shows an approximately $6.99\%$ improvement in mAP metric.
Among all the methods, our method achieves the best performance on mAP, NDCG, and ANMRR, and the numerical difference with the optimal method on NN is comparable. Specifically, our method achieves the gains of $11.93\%\text{-}99.21\%, 2.17\%\text{-}9.52\%$ and $2.18\%\text{-}42.93\%$ in terms of mAP, NN and DCG, respectively, and the decline of $33.67\%\text{-}44.04\%$ in ANMRR compared to TT\_VODKA, CU\_MM, MM\_AI\_SoCSE\_KLET, NUC\_AICV and Ome\_Candy approaches. It can be said that our method is effective for 3d model retrieval tasks in the open-set domain.

\begin{table}[H]
\centering
\caption{The comparison of actually used modality for each method.}
\label{Tab.2}
\resizebox{\linewidth}{!}{
\begin{tabular}{ccccc}
\hline
Method & Mutil-view & Pointcloud & Mesh & Voxel  \\ 
\hline
 TT\_VODKA\cite{feng2022shrec} & \Checkmark & \XSolidBrush & \XSolidBrush & \Checkmark \\ 
 CU\_MM\cite{feng2022shrec} & \Checkmark & \Checkmark & \XSolidBrush & \XSolidBrush \\ 
 MM\_AI\_SoCSE\_KLETech\cite{feng2022shrec} & \Checkmark & \Checkmark & \XSolidBrush & \XSolidBrush \\
 NUC\_AICV\cite{feng2022shrec} & \Checkmark & \Checkmark & \Checkmark & \Checkmark \\
 Ome\_Candy\cite{feng2022shrec} & \Checkmark & \XSolidBrush & \XSolidBrush & \XSolidBrush \\
 HCMUS\cite{feng2022shrec} & \Checkmark & \Checkmark & \Checkmark & \Checkmark \\
 Ours  & \Checkmark & \Checkmark & \Checkmark & \Checkmark \\
\hline
\end{tabular}
}
\end{table}

\begin{table}[H]
\centering
\caption{Evaluation results on OS-MN40 and OS-MN40-Miss.}
\label{Tab.3} 
\resizebox{\linewidth}{!}{
\begin{tabular}{cccccc}
\hline
Dataset & Method & mAP & NN & NDCG & ANMRR  \\ 
\hline
 &TT\_VODKA & 0.3293 & 0.8094 & 0.5017 & 0.6519 \\ 
 &CU\_MM & 0.4083 & 0.8677 & 0.5773 & 0.5836 \\ 
 &MM\_AI\_SoCSE\_KLETech & 0.4152 & 0.8646 & 0.5808 & 0.5750 \\
 OS-MN40 &NUC\_AICV& 0.4183 & 0.8635 & 0.5660 & 0.5750 \\
 &Ome\_Candy & 0.445 & 0.8083 & 0.5869 & 0.5500 \\
 &HCMUS & 0.5861 & \textbf{0.9240} & 0.7018 & 0.4308 \\
 &Ours  & \textbf{0.6560} & 0.8865 & \textbf{0.7171} & \textbf{0.3648} \\
\hline
&HCMUS & 0.2873 & 0.7885 & 0.4632 & 0.6789 \\
OS-MN40&TT\_VODKA & 0.2976 & 0.7823 & 0.4710 & 0.6809 \\
-Miss&MM\_AI\_SoCSE\_KLETech & 0.3893 & 0.8323 & 0.5489 & 0.5937 \\
&Ours& \textbf{0.4385} & \textbf{0.8667} & \textbf{0.5937} & \textbf{0.5542} \\
\hline
\end{tabular}}
\end{table}

For the OS-MN40-Miss, we replace the missing modality features with zero vectors, which is also done by other methods using either zeros or random numbers. Overall, our method outperforms other methods on all four evaluation metrics. However, it is worth noting that the performance decreases significantly (about $20\%$) when modalities are randomly dropped.

\begin{figure}[htpb]
\centering
\includegraphics[scale=0.5]{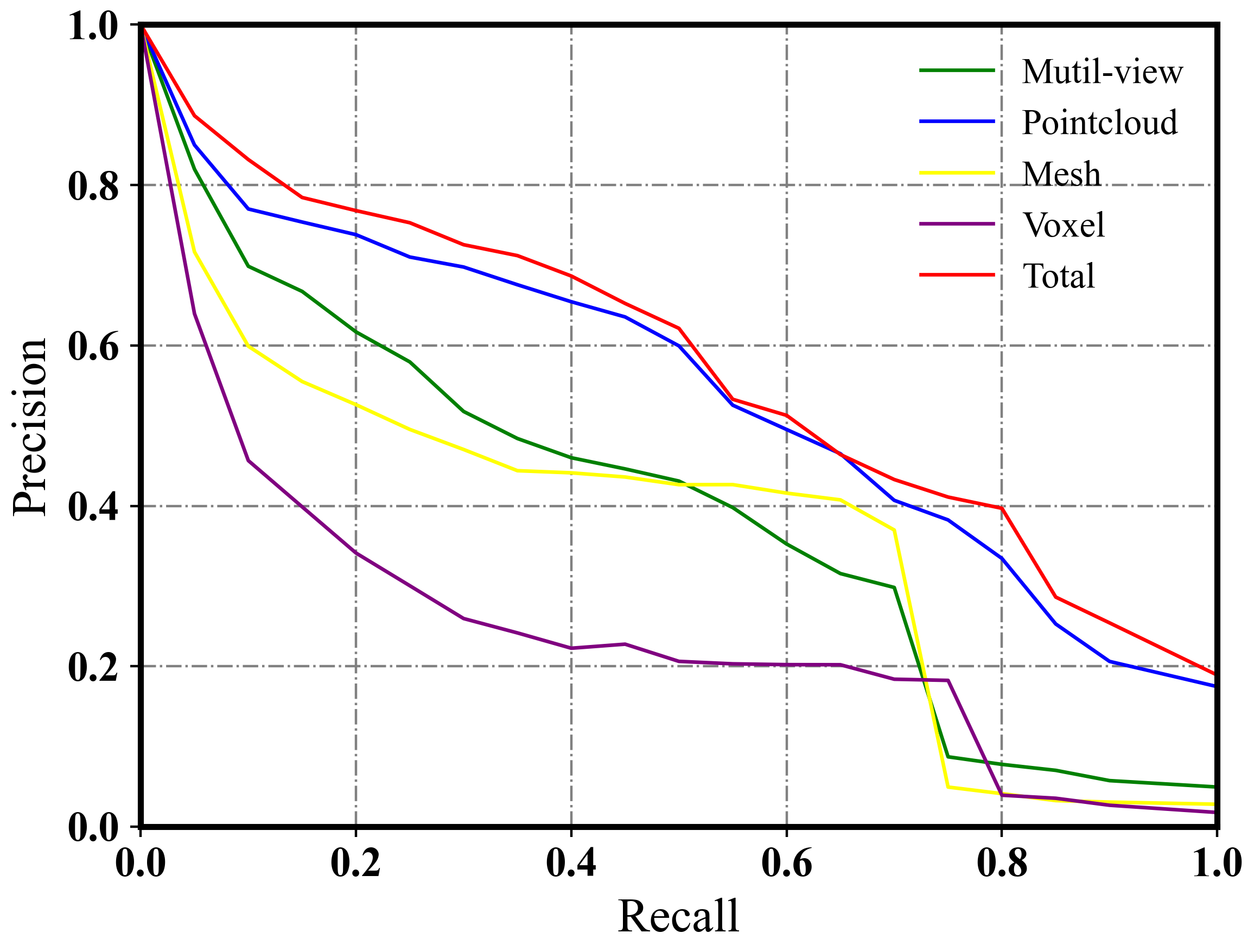}
\caption{The PR Curve of different modalities of our method.}
\label{fig.4}
\end{figure}

In order to further observe the performance of different modalities in retrieval tasks, PR curves of different modalities were drawn based on our method, as shown in Fig.\ref{fig.4}. It can be seen that the performance of the four modalities of multi-view, point cloud, mesh, and voxel on the retrieval task varies greatly in our method. Point cloud performs best in these four modalities, followed by Multi-view representation, mesh, and voxel. In addition, the multi-modal information fusion method we use is more efficient than the method using only a single modality.


\subsection{Ablation Study}


In Tab.\ref{Tababl} we report the results of an ablation study, obtained by removing each core component of our method in isolation, i.e., the self-supervised correlated loss and the TD algorithm correlated loss. We conducted a study on the accuracy of novel class discovery on the OS-MN40 dataset. Additionally, since the majority of data in the OS-MN40 is obtained from ModelNet40, we also report the classification accuracy of our model on the ModelNet40 dataset under supervised conditions to further evaluate its performance.

\begin{table}[H]
\centering
\caption{The accuracy of NCD task on OS-MN40 and supervised classification task of ModelNet40.} 
\label{Tab.4}
\begin{tabular}{ccc}
\hline
    Loss    &\multicolumn{2}{c}{Dataset}  \\ \cmidrule(r){2-3}
    Function & OS-MN40  & ModelNet40 \\             
\hline
$\mathcal{L}_{CE}$ & 0.752 & 0.986   \\
$\mathcal{L}_{CE}+\mathcal{L}_{TD}$ & 0.769 & 0.979  \\
$\mathcal{L}_{CE}+\mathcal{L}_{SS}$ & 0.763 & 0.935   \\
$\mathcal{L}_{CE}+\mathcal{L}_{TD}+\mathcal{L}_{SS}$ & 0.773 & 0.926  \\
\hline
\end{tabular}
\end{table}

In terms of supervised classification accuracy, our multi-modal fusion network demonstrates excellent performance, validating its effectiveness in 3D model classification tasks. We found that using the cross-entropy loss alone achieves the highest accuracy in supervised learning while incorporating self-supervised learning and reinforcement learning slightly decreases the accuracy. However, in the task of novel class discovery, the inclusion of $\mathcal{L}_{SS}$ and $\mathcal{L}_{TD}$ has improved classification accuracy by about $2\%$. This indicates that under open-set conditions, the integration of self-supervised learning and reinforcement learning with rewards effectively enhances the model's performance. Nonetheless, in a supervised situation, the use of overly complex loss functions may hinder the model's performance.


\begin{table*}[htpb]
\centering
\caption{The evaluation of mAP metric for the usage of different components at different epoch counts on OS-MN40.} 
\label{Tababl}
\vspace{0.9em} \centering
\begin{tabular}{ccccccccc}
\hline
Cross-Entropy & TD algorithm & Self-Supervision & STLClu & 5 epochs & 10 epochs & 20 epochs & 30 epochs & 40 epochs  \\ 
\hline
\Checkmark & \XSolidBrush & \XSolidBrush & \XSolidBrush 
& 0.3425 & 0.3413 & 0.3396 & 0.3422 & 0.3408 \\ 
\Checkmark & \XSolidBrush & \XSolidBrush & \Checkmark
& 0.3812 & 0.5549 & 0.5886 & 0.5936 & 0.5962 \\ 
 \Checkmark & \Checkmark & \XSolidBrush & \XSolidBrush 
 & 0.3461 & 0.3414 & 0.3423 & 0.3443 & 0.3425 \\ 
 \Checkmark & \Checkmark & \XSolidBrush & \Checkmark
 & 0.4064 & 0.5987 & 0.6354 & 0.6409 & 0.6423 \\ 
 \Checkmark & \XSolidBrush & \Checkmark & \XSolidBrush 
 & 0.3524 & 0.3613 & 0.3764 & 0.3696 & 0.3802 \\ 
 \Checkmark & \XSolidBrush & \Checkmark & \Checkmark 
 & 0.3802 & 0.5618 & 0.6102 & 0.6188 & 0.6203 \\ 
\Checkmark & \Checkmark & \Checkmark & \XSolidBrush 
& 0.3511 & 0.3645 & 0.3782 & 0.3801 & 0.3799 \\ 
\Checkmark & \Checkmark & \Checkmark & \Checkmark 
& 0.4152 & 0.6072 & 0.6488 & 0.6560 & 0.6560 \\ 
\hline
\end{tabular}
\end{table*}

\begin{table*}[htpb]
\centering
\caption{Evaluation of the mAP metric for different modalities used on OS-MN40.} 
\label{Tabmulti}
\begin{tabular}{cccccccccccc}
\hline
S.N. & Multi-view & Pointcloud & Mesh & Voxel & mAP & S.N. & Multi-view & Pointcloud & Mesh & Voxel & mAP  \\ \cmidrule(r){1-6}  \cmidrule(r){7-12}

1 & \Checkmark   & \XSolidBrush & \XSolidBrush & \XSolidBrush & 0.3895 & 9 & \XSolidBrush & \Checkmark   & \XSolidBrush & \Checkmark   & 0.5250 \\ 
2 & \XSolidBrush & \Checkmark   & \XSolidBrush & \XSolidBrush & 0.4376 & 10& \XSolidBrush & \XSolidBrush & \Checkmark   & \Checkmark   & 0.5325 \\ 
3 & \XSolidBrush & \XSolidBrush & \Checkmark   & \XSolidBrush & 0.4102 & 11& \Checkmark   & \Checkmark   & \Checkmark   & \XSolidBrush & 0.6088 \\ 
4 & \XSolidBrush & \XSolidBrush & \XSolidBrush & \Checkmark   & 0.3545 & 12& \Checkmark   & \Checkmark   & \XSolidBrush & \Checkmark   & 0.6132 \\ 
5 & \Checkmark   & \Checkmark   & \XSolidBrush & \XSolidBrush & 0.5143 & 13& \Checkmark   & \XSolidBrush & \Checkmark   & \Checkmark   & 0.5859 \\ 
6 & \Checkmark   & \XSolidBrush & \Checkmark   & \XSolidBrush & 0.4712 & 14& \XSolidBrush & \Checkmark   & \Checkmark   & \Checkmark   & 0.5609 \\
7 & \Checkmark   & \XSolidBrush & \XSolidBrush & \Checkmark   & 0.4509 & 15& \Checkmark   & \Checkmark   & \Checkmark   & \Checkmark   & 0.6423 \\ 
8 & \XSolidBrush & \Checkmark   & \Checkmark   & \XSolidBrush & 0.5497 &   &   &   &   &   &   \\ 
\hline
\end{tabular}
\end{table*}

\begin{figure}[H]
\centering
\includegraphics[scale=0.5]{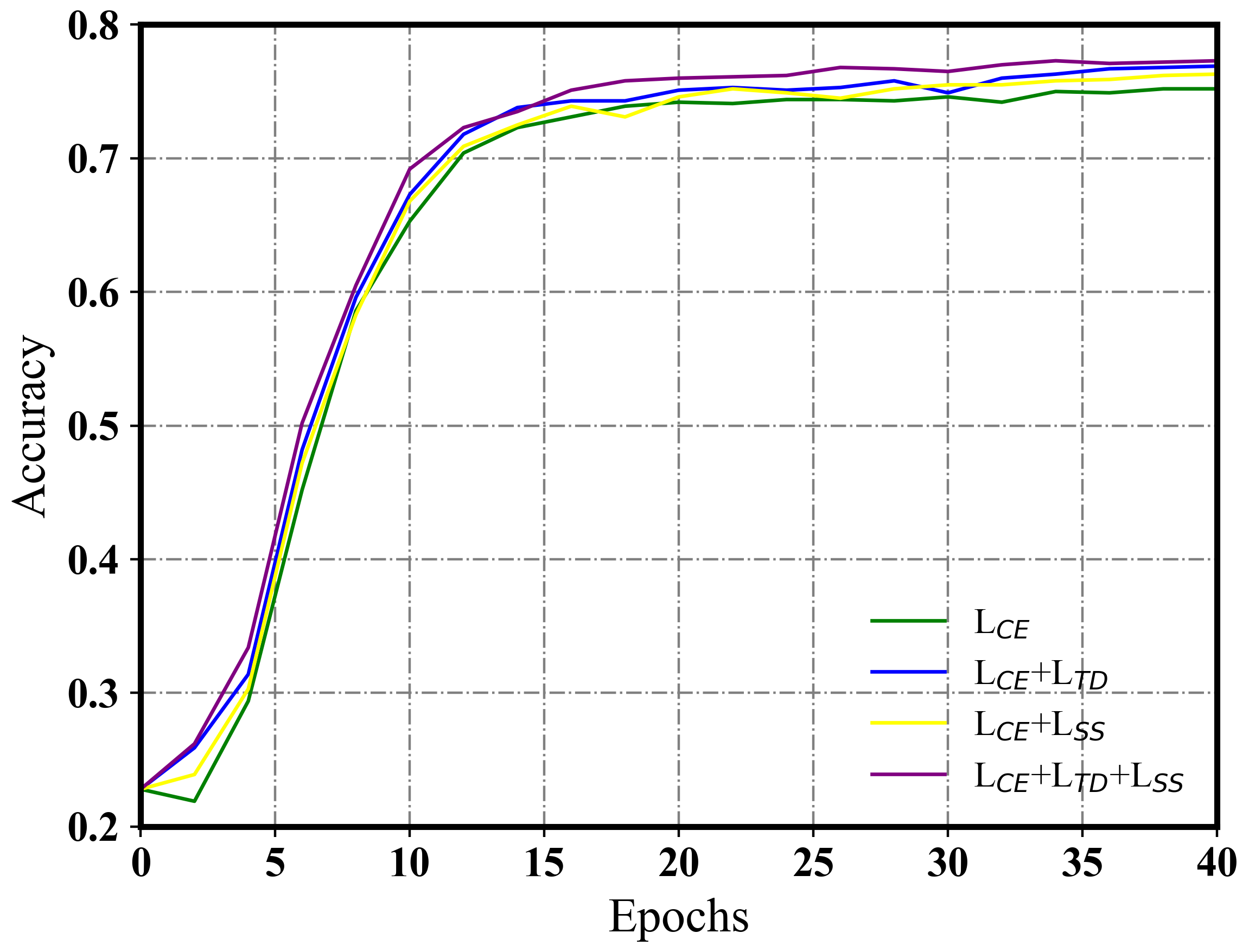}
\caption{The relationship between classification accuracy and epochs with different components}
\label{fig.51}
\end{figure}

Additionally, we examined the correlation between classification accuracy and epochs by utilizing different components. Detailed results can be found in Fig.\ref{fig.51}. The figure clearly demonstrates that incorporating reinforcement learning and self-supervised learning enhances the model's training performance by about $2\%$-$4\%$.

To further investigate the influence of different components on model training, we conducted an evaluation of the mAP metric using various components at different iteration counts on the OS-MN40 dataset. 
The results are presented in Tab.\ref{Tababl}. The assessed components encompassed the self-supervised loss function, TD algorithm, and STLClu strategy. Based on the findings from the table, the following conclusions can be drawn:
\begin{itemize}
\item[$\bullet$]
The STLClu strategy constitutes the most essential component of our method. Without utilizing STLClu, our method struggles to enhance the retrieval performance for 3D models during the training process. When STLClu is omitted for generating labels for unlabeled data, the incorporation of self-supervised learning leads to an approximate $4\%$ improvement in the model's mAP metric.
\item[$\bullet$]
The integration of self-supervised loss and TD algorithm significantly enhances the model's performance throughout the training procedure. Specifically, the inclusion of $\mathcal{L}_{SS}$ leads to a $3\%$ improvement in the retrieval task, while $\mathcal{L}_{TD}$ contributes to a $5\%$ improvement. When combined, these two components yield a cumulative performance gain of $6\%$ for the model.
\end{itemize}

In the previous section, we conducted comparative experiments with other open-set 3D model retrieval algorithms and analyzed their retrieval performance, including a comparison of algorithms using different modalities of data as input. Based on the comparison experiments, we can conclude that incorporating additional modalities as input can effectively enhance the performance of the network model. However, due to variations in network structures, training strategies, and utilization of input information across different methods, the full validation of the impact of different modalities on network model performance is challenging. To thoroughly evaluate the representational capacity of different modalities in 3D models and their influence on network model performance, we conducted ablation experiments on network performance with different modalities. To ensure controlled variables, these experiments were performed on our proposed method using identical parameters and training procedures. Additionally, since self-supervised learning involves contrasting multi-modal data, this component was excluded from experiments. In our feature fusion network, the feature of missing modalities is set to zero. The results of the ablation experiments for different modalities are presented in Tab.\ref{Tabmulti}, from which we can derive the following conclusions:
\begin{figure*}
\centering
\includegraphics[scale=0.73]{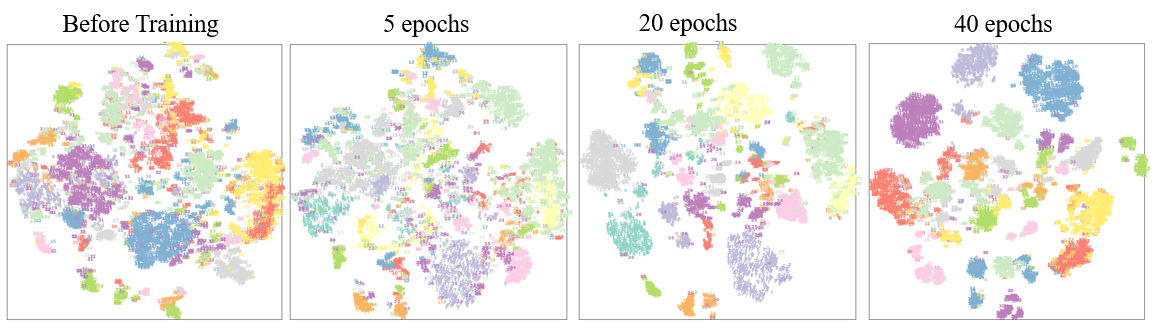}
\caption{T-SNE visualization of unlabeled samples on OS-MN40. As the training progresses, data of the same category gradually cluster together, while data of different categories progressively separate.}
\label{fig.5}
\end{figure*}

\begin{itemize}
\item[$\bullet$]
Within our method's framework, point-cloud demonstrates the highest performance among the four single-modal data types, followed by mesh, multi-view, and voxel, which exhibit comparatively lower performance.
\item[$\bullet$]
When incorporating two modalities, the model's performance further improves compared to using a single modality. However, upon comparing the experimental results of S.N.2, S.N.3, S.N.9, and S.N.10, we observed that point-cloud outperforms mesh in the single-modal case. Nevertheless, after integrating voxel information, point-cloud combined with voxel performs worse than mesh combined with voxel, indicating redundancy between different modalities.
\item[$\bullet$]
The average mAP scores for utilizing single-modal, dual-modal, triple-modal, and all four modalities inputs are 0.3980, 0.5071, 0.5922, and 0.6423, respectively. With an increasing number of input modalities, the network model's performance exhibits an upward trend, suggesting that multi-modal information contributes to a more comprehensive understanding of the data by the network model.
\end{itemize}

\subsection{Visualization Experiment}
In addition to quantitative results, we also report a qualitative analysis showing the feature space learned by our method on OS-MN40.
We use PCA\cite{abdi2010principal} to reduce the dimensionality of the fused global feature vectors $l_i$ for all unlabeled data, and then we project the data in two dimensions using t-SNE\cite{van2014accelerating}. 

As shown in Fig.\ref{fig.5}, we can observe that after retraining the model with labeled data, it has already begun to aggregate data of the same categories to a certain extent. However, there are still samples where data of different categories are clustered together. After 5 epochs, we observe that data of the same categories have become somewhat dispersed, but the distinction between data of different categories has improved compared to before. This observation validates the correctness of our STLClu strategy, which necessitates progressively relaxing the constraints during training. It can be seen that after 40 epochs, samples of different categories in the unlabeled dataset become more distinguishable, and samples of the same categories are more similar. This shows that our approach enables neural networks to perform better when dealing with Novel Class Discovery problems in the open-set.

\begin{figure}[H]
\centering
\includegraphics[scale=0.58]{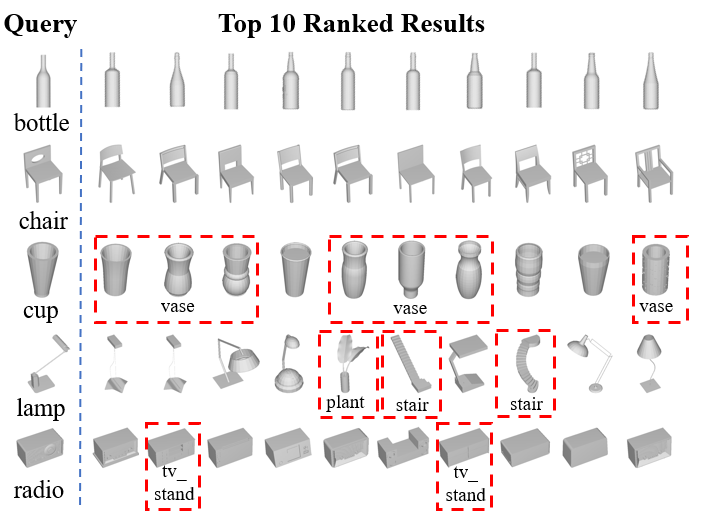}
\caption{Retrieval examples using the proposed method on OS-MN40. The first column presents the queries, while the subsequent columns display the ranked retrieval results. A red dotted line box is used to indicate instances where an error occurred during the retrieval of a 3D model.}
\label{fig.6}
\end{figure}

In order to evaluate the performance of our method in an open-set scenario, we conducted a random selection of five distinct categories of 3D models from the query dataset. The top ten retrieval results for each category were visualized in Fig.\ref{fig.6}. We made the following observations:

\begin{itemize}
\item[$\bullet$]
Retrieval results for bottles and chairs exhibited relatively good performance, whereas lamps and radios showed acceptable retrieval results. Notably, cups performed the worst among all the categories.
\item[$\bullet$]
Categories that achieved better retrieval performance are more abundant in the dataset, enabling them to benefit from more effective training.

\item[$\bullet$]
Cups exhibited poor retrieval performance, with a mere $0.8\%$ representation in the training dataset. Interestingly, vases, which bear a striking resemblance to cups, had representation nine times greater. Upon scrutinizing the retrieval results, discerning the disparity between the retrieved results and the target objects proved challenging for human evaluators. This implies that our method encounters difficulty in achieving competitive performance in the retrieval task when confronted with limited training data.
\item[$\bullet$]
Overall, our method demonstrates the capability to identify results with a high degree of similarity to the target objects in the retrieval task.
\end{itemize}

\subsection{Transfer Experiment On 2D Image}
The experimental results presented in the preceding subsections have confirmed the effectiveness of our method in addressing the challenge of novel class discovery in 3D models. In order to further substantiate the efficacy of our framework, we endeavored to apply this framework to the novel class discovery problem in the field of 2D images.

\begin{table}[H]
\centering
\caption{Comparison with state-of-the-art methods on CIFAR-10 for novel class discovery.} 
\label{Tab.7}
\begin{tabular}{cccc}
\hline
      &\multicolumn{3}{c}{Classification Accuracy}  \\ \cmidrule(r){2-4}
   Method & Labeled & Novel & All\\
\hline
DTC\cite{han2019learning}     & 0.539 & 0.395 & 0.383   \\
CGDL\cite{9157396} & 0.723 & 0.446 & 0.397  \\
UNO\cite{fini2021unified}     & 0.916 & 0.693 & 0.805   \\
ORCA\cite{cao2022openworld}         &0.882 &0.904 & 0.897\\
OpenLDN\_UDA\cite{rizve2022openldn} & 0.957 & 0.951 & 0.954 \\
Ours    & \textbf{0.961} & \textbf{0.952} & \textbf{0.956}  \\
\hline
\end{tabular}
\end{table}


We compared our approach with state-of-the-art approaches and presented the classification accuracy of each method on the labeled data, unlabeled data with novel classes, and all data in the dataset. The detailed results are provided in Tab.\ref{Tab.7}.
The analysis shows that our approach achieves state-of-the-art performance on novel class discovery problems in 2D images. This observation indicates the potential benefits of the proposed network architecture for research in both the open-set 3D and 2D domains.

\section{Conclusions}

In this paper, we introduce a novel Reinforcement Learning Multi-modal Feature Fusion Network to address the challenge of open-set information classification and retrieval. Our approach involves utilizing the STLClu strategy to gradually assign labels to unlabeled data during the training procedure. We adopt the MLMA architecture for processing and integrating multi-modal data, enabling the incorporation of self-supervised learning techniques for enhanced model training. Additionally, we optimize the agents' classification strategies using rewards derived from environmental feedback.
These strategic choices enhance the training paradigm of our model, aligning it more closely with human learning processes and yielding a more intelligent model. To substantiate the efficacy of our methodology, we conduct extensive experiments on the OS-MN40 and OS-MN40-Miss datasets, achieving competitive results.

\section*{Acknowledgments}
This work was supported in part by the National Natural Science Foundation of China (62272337,62072232) and the Natural Science Foundation of Tianjin (16JCZDJC31100, 16JCZDJC31100).

\bibliographystyle{unsrt}
\bibliography{ref}

\end{document}